\documentclass[11pt]{article}

\usepackage[final]{acl}
\usepackage{booktabs}
\usepackage{multirow}
\usepackage{amsmath}
\usepackage{times}
\usepackage{latexsym}
\usepackage{graphicx}
\usepackage{amssymb}
\usepackage[T1]{fontenc}

\usepackage[utf8]{inputenc}

\usepackage{microtype}

\usepackage{inconsolata}

\usepackage{graphicx}

\title{ActTraitBench: Quantifying the Knowledge--Decision Gap in Large Language Models via Human-Grounded Behavioral Validation}

\author{
    \textbf{Yutong Yang\textsuperscript{1}} \quad
    \textbf{Chenxi Miao\textsuperscript{2}} \quad
    \textbf{Weikang Li\textsuperscript{2,*}} \quad
    \textbf{Yunfang Wu\textsuperscript{1,*}} \\
    \textsuperscript{1}School of Computer Science, Peking University \\
    \textsuperscript{2}Baidu Inc. \\
    \texttt{yytpku@pku.edu.cn, miaochenxi@baidu.com} \\
    \texttt{wavejkd@pku.edu.cn, wuyf@pku.edu.cn}
}

\begin{document}

\maketitle
\let\thefootnote\relax\footnotetext{\textsuperscript{*}Corresponding authors.}
\begin{abstract}
While Large Language Models (LLMs) can convincingly simulate personas in explicit self-reports, they often deviate in implicit behavioral decisions, revealing a substantial Knowledge--Decision Gap ($G_{\text{KD}}$). Existing benchmarks struggle to measure this discrepancy due to limited construct validity, multidimensional entanglement, and distributional biases in LLM-based evaluation. To address these issues, we propose ActTraitBench, a human-grounded evaluation framework for measuring personality consistency in LLMs. Grounded in empirical human data, ActTraitBench establishes one-to-one mappings between psychometric facets and behavioral paradigms and applies Distributional Calibration via Quantile Mapping to reduce distributional mismatch between LLM-judge scores and human responses. Experiments on 14 mainstream LLMs reveal substantial knowledge--decision gaps and show that assigned personas are reflected more consistently in self-reports than in behavioral decisions for most evaluated models. To mitigate this gap, we further introduce the Chain of Cognitive Alignment (CoCA), an inference-time intervention that reduces $G_{\text{KD}}$ for 12 of the 13 models with paired results. Code and resources are available at \url{https://github.com/Selina233/ActTraitBench}.
\end{abstract}

\section{Introduction}
\label{sec:intro}

As Large Language Models (LLMs) are increasingly deployed in real-world interactive and decision-making scenarios, accurately measuring their emergent psychological traits has become a critical necessity for trustworthy AI. These latent behavioral tendencies directly influence user trust, signal potential alignment risks such as sycophancy, and dictate the functional reliability of autonomous agents \cite{Han2025ThePI}.
Early psychometric endeavors primarily mirrored traditional psychological methodologies, deploying explicit self-report questionnaires (e.g., the Big Five Inventory \cite{soto2017next} or Hexaco
questionnaire \cite{ashton2009hexaco}) directly to LLMs to infer their trait distributions \cite{miotto2022gpt,jiang2023evaluating,serapio2023personality}.
As evaluation paradigms evolved, researchers challenged the overall validity of explicit self-reports, exposing systemic vulnerabilities such as limited temporal stability and artificially highlighted prosociality \cite{bodrovza2024personality, suhr2025challenging}. This catalyzed a transition toward more ecologically valid behavioral measurements, including dynamic multi-agent discourse \cite{bhandari2025can}, contextualized action-selection scenarios \cite{shen-etal-2025-mind}, and experimental psychological paradigms \cite{Han2025ThePI}. 

Recent studies have identified a ``personality illusion'' and ``Value-Action Gap'': LLMs may follow explicit persona descriptions in self-reports while exhibiting less consistent personality-related behavior in contextualized decisions.

Despite this progress, current implicit behavioral evaluation frameworks still suffer from three critical methodological limitations: 

(1) \textbf{Unverified Measurement Validity}: Existing frameworks frequently rely on synthetically generated behavioral labels \cite{shen-etal-2025-mind} or LLM-based trait extraction \cite{bhandari2025can} without strict empirical human benchmarking. This leaves it unverified whether a specific downstream text or decision genuinely and uniquely maps to a latent psychological state. 

(2) \textbf{Multi-dimensional Entanglement}: Complex interactive scenarios, such as multi-agent debates \cite{bhandari2025can} or macro-behavioral tasks \cite{Han2025ThePI}, implicitly force models to process multiple macro-traits (e.g., the entire Big Five) simultaneously. This infuses behavioral scores with high-dimensional confounding noise, making it nearly impossible to isolate the influence of specific psychometric facets. 

(3) \textbf{Distributional Bias}: Current evaluations may also exhibit uncalibrated distributional shifts. For instance, models frequently exhibit default compliance biases \cite{shen-etal-2025-mind}, while LLM-as-a-judge evaluators systematically avoid predicting negative traits like Neuroticism \cite{bhandari2025can}. These biases can introduce systematic mismatches with empirical human distributions.

To address these methodological bottlenecks, we present \textbf{ActTraitBench}, a human-grounded evaluation framework designed to quantify the \textbf{``Knowledge--Decision Gap''} ($G_{\text{KD}}$) in LLMs—the discrepancy between a model's explicit personality self-reports and its behavioral decisions.

Our contributions are three-fold:

(1) \textbf{Human-Grounded Benchmark \& Calibration}: We introduce ActTraitBench, featuring 11 micro-situational paradigms directly mapped to Big Five psychometric facets. By grounding these tasks in empirical human data and applying a non-parametric Quantile Mapping mechanism, we provide human-grounded evidence for construct validity and reduce distributional mismatch between LLM-judge scores and human responses.

(2) \textbf{Quantification of the Knowledge--Decision Gap}: Across 14 mainstream models, we quantify the Knowledge--Decision Gap. Our results show that knowledge--decision gaps persist across the evaluated model generations, suggesting that improvements in model scale and capability do not necessarily lead to better alignment between explicit personality reports and behavioral decisions.

(3) \textbf{Inference-Time Mitigation}: We propose the Chain of Cognitive Alignment (CoCA), an inference-time intervention for reducing the knowledge--decision gap between personality self-reports and behavioral decisions.


\section{Related Work}
\label{sec:related_work}

\subsection{Psychometric Evaluations and Their Limitations}

Early assessments of LLM personality primarily relied on standardized human self-report inventories, prompting models to complete Likert scales directly to maintain alignment with conventional human psychometric protocols \cite{miotto2022gpt,jiang2023evaluating}. Studies have demonstrated that instruction-tuned models can generate personality distributions resembling those of human samples, with larger models exhibiting higher internal consistency and construct validity \cite{pellert2024ai,miotto2022gpt,jiang2023evaluating,serapio2023personality}.

However, this mapping approach has faced significant scrutiny due to fundamental psychometric flaws, such as the lack of measurement invariance and structural validity \cite{suhr2025challenging}. LLMs frequently exhibit systemic ``acquiescence bias,'' providing affirmative responses to conflicting items and rendering traditional reliability metrics like Cronbach's alpha potentially misleading \cite{suhr2025challenging}. Furthermore, research on temporal stability indicates that LLM persona representations lack consistency, with flagship models often demonstrating poor test-retest reliability \cite{bodrovza2024personality}. These findings suggest that static self-reports are insufficient for capturing robust behavioral tendencies.

\begin{figure*}[t]
\centering
\includegraphics[width=0.95\linewidth]{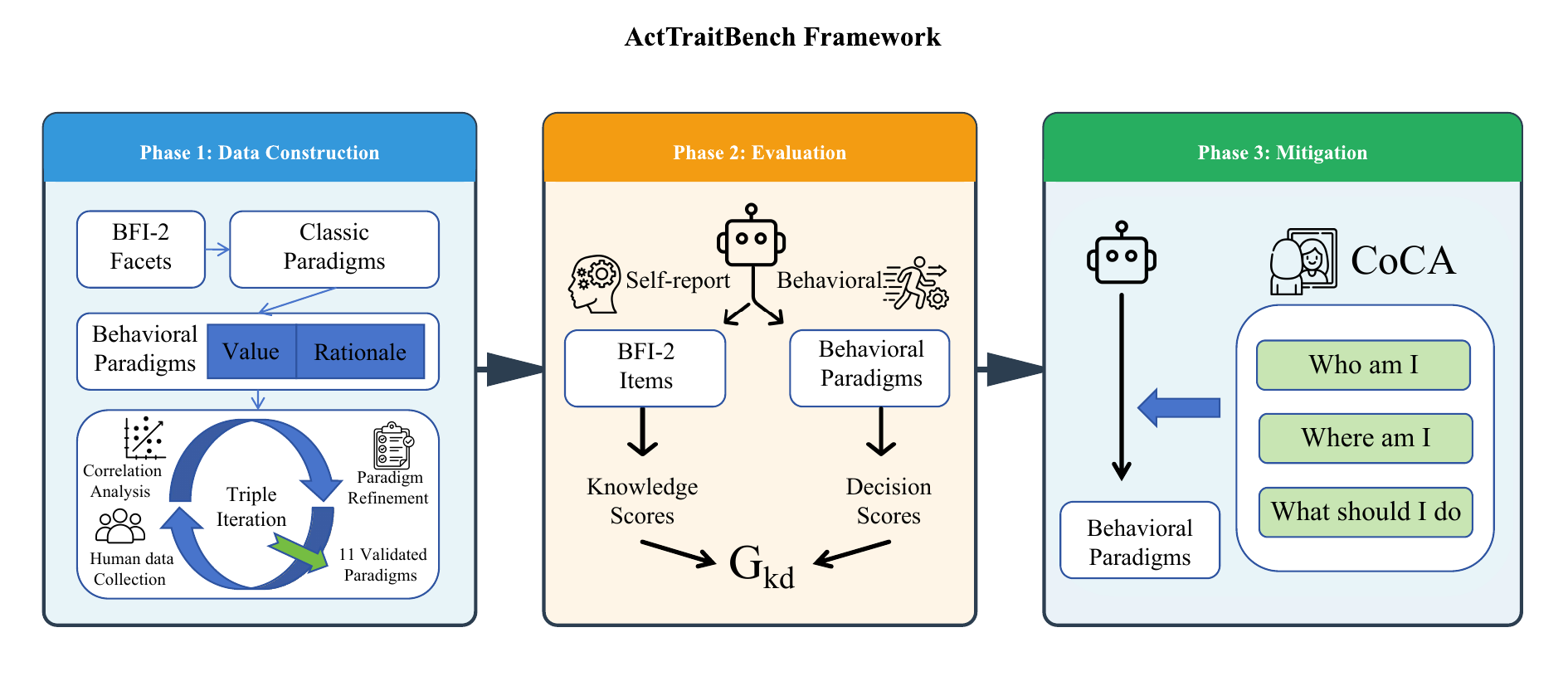}
\caption{The overall architecture of the ActTraitBench framework. 
}
\label{fig:overall_pipeline}
\end{figure*}

\subsection{Implicit Behavioral Assessment and the Knowledge--Decision Gap}

To overcome these limitations, recent literature has shifted toward behavioral assessment paradigms that examine model tendencies through interactive behaviors or contextualized tasks. For instance, studies have revealed that LLMs suffer from ``persona drift'' in multi-turn discourse, where they regress toward neutral patterns \cite{bhandari2025can}, or exhibit a significant ``value-action gap'' when comparing stated values to decision-making actions across validated scenarios \cite{shen-etal-2025-mind}. 

Furthermore, researchers have explored the ``personality illusion'' by deploying established experimental psychological tasks to observe behavioral shifts under trait manipulations \cite{Han2025ThePI}. However, these paradigms face methodological hurdles: from a psychometric perspective, classical behavioral tasks often recruit multiple latent dimensions simultaneously, potentially entangling factors such as risk preference with diverse psychometric facets. This high-dimensional confounding makes it difficult to disentangle the contribution of specific traits, raising concerns regarding the construct validity of such implicit measures. Task context introduces an additional challenge for behavioral personality assessment. Zou et al. find that the relationship between LLM self-reported personality and human-perceived personality varies across interaction tasks \cite{zou2025can}, making it difficult to determine whether an expected trait is absent because the model fails to translate personality into behavior or because the task provides limited opportunity for that trait to be expressed. ActTraitBench addresses this ambiguity by using facet-level behavioral scenarios that are retained only when they exhibit the expected trait--behavior associations in human participants, providing a more controlled basis for evaluating personality-to-behavior consistency.

\section{Method}
\label{sec:method}

Figure~\ref{fig:overall_pipeline} provides an overview of the \textbf{ActTraitBench} framework, which systematically encompasses three core phases: paradigm development through iterative human-grounded refinement, quantitative evaluation of the Knowledge--Decision Gap ($G_{\text{KD}}$), and the mitigation process using the Chain of Cognitive Alignment (CoCA) strategy.

\subsection{Data Construction Pipeline: A Triple-Iteration Loop}
\label{subsec:data_pipeline}

To ensure that the implicit behavioral tests accurately and objectively reflect the underlying personality traits of Large Language Models (LLMs), we establish a human-data-driven pipeline characterized by a triple-iteration loop for item refinement.

\paragraph{Paradigm Design and Lewin's Equation}
The initial experimental scenarios are meticulously designed to map onto the granular facets of the BFI-2 framework \cite{john1999big}. To guarantee psychometric validity, all paradigms are developed by researchers with dual academic training in both computer science and psychology, adapted from classic paradigms in experimental psychology \cite{schachter1959psychology,asch2016effects,ariely2002procrastination} and behavioral economics \cite{dhar2000trying,schachter1959psychology,eliaz2010paying}. In formulating these interactive scenarios, we adhere strictly to the foundational psychological principle of \textbf{Lewin's Equation} \cite{Lewin1936PrinciplesOT}, which posits that behavioral outcomes are not isolated linear outputs of personality traits, but rather the reconstructed products of underlying interactions between the person ($P$) and their immediate environment ($E$):
\begin{equation}
    B = f(P, E)
\end{equation}
Consequently, to isolate and extract the persona variable $P$ precisely within our experiments, the environmental variable $E$ must be rigorously controlled. Scenarios are engineered to be as parsimonious, straightforward, and clear as possible, minimizing extra cognitive noise introduced by complex task environments.

\paragraph{Two-stage Prompting Protocol}
To systematically dissect the latent motivations behind a model's choices, each test item leverages a ``Two-stage'' interaction flow:
\begin{itemize}
    \item \textbf{Stage 1 (Quantitative Decision)}: The subject is required to input a numerical value within a continuous scale of 0--99 (e.g., points allocated, willingness-to-pay premium), providing a precise quantification of their explicit behavioral inclination.
    \item \textbf{Stage 2 (Qualitative Attribution)}: Triggered only after the completion of Stage 1, this follow-up prompt requires the subject to state the underlying logic and rationale behind their numerical decision (approximately 50 words).
\end{itemize}
We use sequential API calls to fetch quantitative scores and qualitative reasoning separately, maintaining strict symmetry with human testing protocols. An implicit behavioral paradigm example is provided in Appendix \ref{sec:appendix_example}.


\paragraph{Human Data Collection and Iterative Refinement}
We developed an interactive Web-based survey system to collect authentic behavioral responses from human cohorts. For the scoring phase, we employ \texttt{gpt-5.4} as an LLM-as-a-judge. When evaluating a subject's combined ``numerical decision and textual rationale,'' the judge prompt explicitly includes the complete question context, the definition of the target BFI-2 facet, reference BFI-2 items, and fine-grained multi-level scoring criteria, ultimately mapping the response to a behavioral trait score ranging from 1 to 5.

Upon acquiring the human data, we systematically compute the \textbf{Spearman's rank correlation coefficient} (Spearman's $\rho$) between the implicit behavioral scores and the explicit baseline scores from the standard BFI-2 self-reports. We use BFI-2 self-reports as a construct reference rather than an independent behavioral gold standard: retained paradigms are those that exhibit the expected trait--behavior associations in human participants. Applying these human-validated associations to LLMs therefore involves a human-to-model transfer assumption, which we discuss further in the limitations. Items that fail to reach statistical significance undergo a rigorous case study and expert discussion within the research team. These scenarios are dynamically restructured and redeployed to the Web platform for subsequent rounds of data collection. Due to this adaptive replacement mechanism, the effective human sample size ($N$) varies across items (N=47 or 94). Following three rounds of strict iterative filtering, we finalize 11 core implicit test paradigms that exhibit robust statistically significant correlations with their corresponding personality facets, as detailed in Table~\ref{tab:correlation_results}.

\begin{figure}[t]
    \centering
    \includegraphics[width=\linewidth]{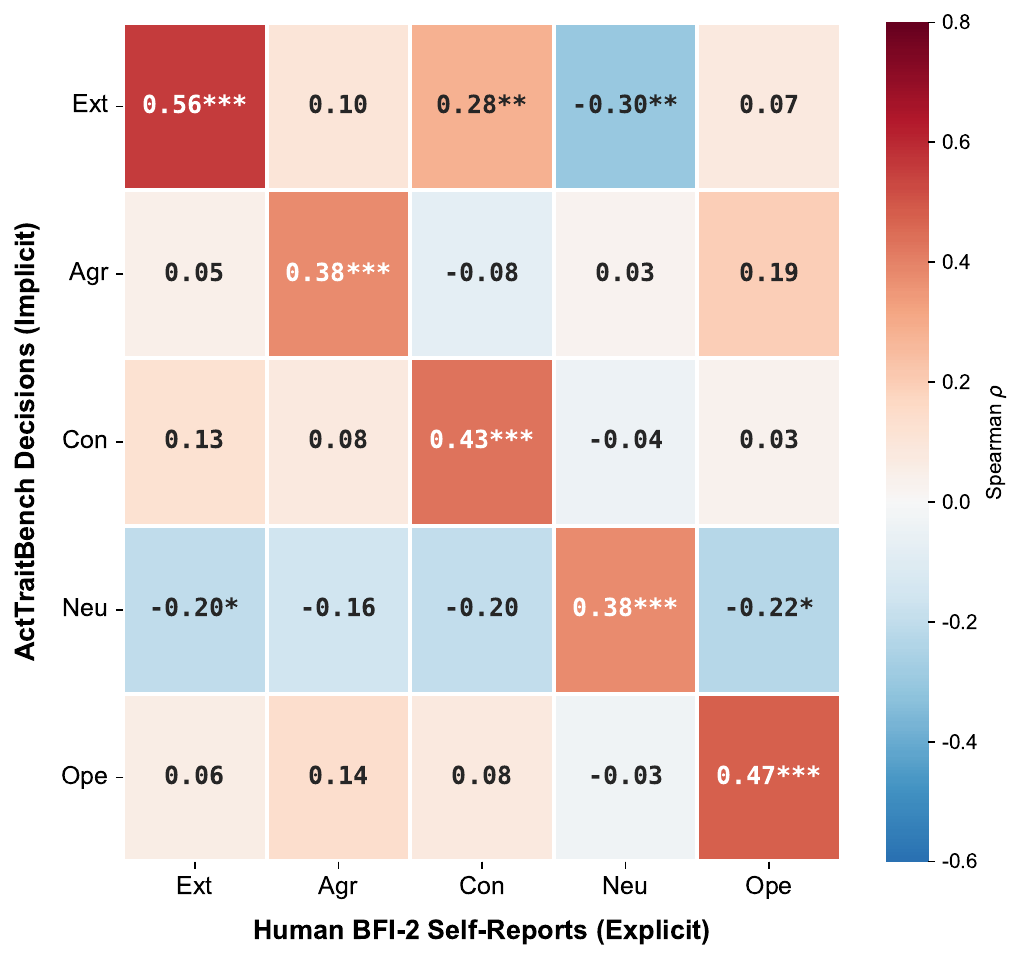}
    \caption{Cross-dimensional correlation matrix between ActTraitBench implicit decisions and explicit BFI-2 self-reports ($N=94$). Values represent Spearman's $\rho$. ${}^*p < 0.05, {}^{**}p < 0.01, {}^{***}p < 0.001$.}
    \label{fig:behavioral_heatmap}
\end{figure}

\begin{figure}[h]
    \centering
    \includegraphics[width=\linewidth]{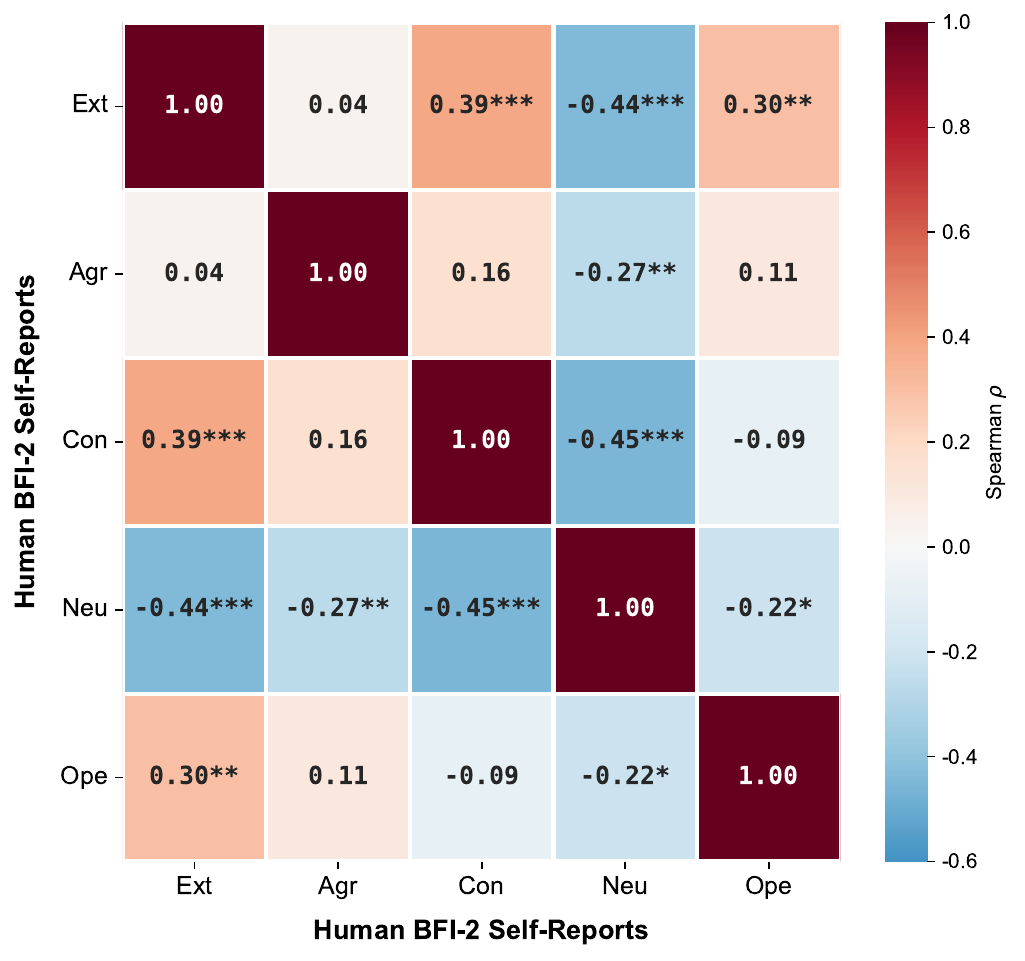}
    \caption{Internal correlation matrix of the empirical human BFI-2 self-report baseline ($N=94$). This baseline illustrates the natural statistical covariation inherently present among the Big Five dimensions in human populations. Values represent Spearman's rank correlation coefficient ($\rho$). ${}^*p < 0.05$, ${}^{**}p < 0.01$, ${}^{***}p < 0.001$.}
    \label{fig:bfi_internal}
\end{figure}

\paragraph{Construct Validity and Dimensional Entanglement}
To evaluate construct validity, we correlated the implicit behavioral decisions with the explicit BFI-2 self-reports (Figure~\ref{fig:behavioral_heatmap}). The results demonstrate robust convergent validity, with all homologous trait pairs along the diagonal exhibiting the highest positive correlations ($0.38 \leq \rho \leq 0.56$, $p < 0.001$). Furthermore, most off-diagonal correlations are low or insignificant, indicating effective mitigation of cross-dimensional interference.

Some cross-dimensional correlations nevertheless remain significant. To examine whether these associations arise from task-induced entanglement or reflect the underlying covariance structure of the Big Five itself, we further compute the internal correlation matrix of the empirical human BFI-2 self-reports (Figure~\ref{fig:bfi_internal}). The human baseline exhibits similar cross-dimensional covariation, including a negative correlation between Extraversion and Neuroticism ($\rho=-0.44$, $p<0.001$) and a positive correlation between Conscientiousness and Extraversion ($\rho=0.39$, $p<0.001$). This correspondence suggests that at least part of the residual cross-dimensional associations observed in ActTraitBench reflects the natural non-orthogonality of the Big Five rather than task-specific confounding. Although our micro-situational design reduces task-induced entanglement, completely isolating a single psychological dimension remains a practical limitation.

\begin{table}[t]
\centering
\small
\begin{tabular*}{\linewidth}{@{\extracolsep{\fill}}lccc@{}}
\toprule
\textbf{Target Facet} & \textbf{$N$} & \textbf{Spearman $\rho$} & \textbf{$p$-value} \\
\midrule
E\_Sociability & 94 & 0.526 & $<0.001$ \\
E\_Assertiveness & 94 & 0.355 & $<0.001$ \\
C\_Productiveness & 94 & 0.543 & $<0.001$ \\
C\_Organization & 47 & 0.411 & 0.004 \\
O\_Curiosity & 94 & 0.356 & $<0.001$ \\
O\_Aesthetic & 94 & 0.406 & $<0.001$ \\
N\_Depression & 94 & 0.325 & 0.001 \\
N\_Volatility & 47 & 0.300 & 0.040 \\
N\_Anxiety & 94 & 0.322 & 0.002 \\
A\_Compassion & 94 & 0.276 & 0.007 \\
A\_Trust & 94 & 0.209 & 0.043 \\
\bottomrule
\end{tabular*}
\caption{Spearman rank correlation analysis of the 11 finalized implicit behavioral paradigms after triple-iteration filtering, evaluated on pre-calibration human data. All retained items demonstrate significant positive correlations ($p < 0.05$) with authentic human BFI-2 self-report scores.}
\label{tab:correlation_results}
\end{table}

\subsection{Distributional Calibration via Quantile Mapping}\label{subsec:calibration}LLM-judges often exhibit biased scoring distributions (e.g., extremity or central clustering). To align these with human BFI-2 norms, we implement a cross-domain mapping function $f: S_{\text{judge}} \to S_{\text{human}}$ using three sequential steps:\paragraph{1. Quantile Extraction}For a raw score $s$, we compute its percentile rank $P$ within the corresponding facet-specific judge-score distribution $\mathcal{D}_{\text{judge}}$:\begin{equation}P = \text{Percentile}(s; \mathcal{D}_{\text{judge}})\end{equation}\paragraph{2. Interval Anchoring}We map $P$ onto the human benchmark distribution $\mathcal{D}_{\text{human}}$. Given an interval $[p_i, p_j]$, we identify the corresponding segment in $\mathcal{D}_{\text{human}}$.\paragraph{3. Score Transformation}The calibrated score $\hat{s}$ is the expectation of human scores within this interval:
\begin{equation}
\resizebox{0.89\hsize}{!}{$
    \hat{s} = \mathbb{E}[S_{\text{human}} \mid \text{Percentile}(S_{\text{human}}; \mathcal{D}_{\text{human}}) \in [p_i, p_j]]
$}
\end{equation}
This approach shifts the evaluation from arbitrary numerical scales to relative positions within the behavioral spectrum, effectively neutralizing systemic bias and ensuring alignment with empirical human distributions.

\subsection{Mitigation Method: Chain of Cognitive Alignment (CoCA)}
\label{subsec:coca}

To mitigate the Knowledge--Decision asymmetry observed in mainstream LLMs, we introduce an inference-time intervention named the Chain of Cognitive Alignment (CoCA).

In experimental psychology, \textbf{Objective Self-Awareness Theory} posits that when individuals engage in self-scrutiny or become aware that their consistency is being evaluated, their actions tend to align more closely with their internal beliefs \cite{duval1972theory,Diener1976EffectsOS}. Drawing inspiration from this framework, CoCA introduces a structured, multi-stage self-reflective mechanism for LLMs. 

Rather than directly eliciting the final task decision, CoCA forces the model to execute a progressive internal reflection protocol structured within a rigid JSON schema prior to action selection. By embedding this metacognitive self-monitoring directly into the model's autoregressive reasoning stream, we enforce a strict cognitive chain. The three progressive reflective tiers within the schema are defined as follows:

\begin{enumerate}
    \item \textbf{Identity Recognition (Who am I?)}: Formulated as \texttt{step\_1} in the reflection block, the model is prompted to explicitly self-assess its assigned psychological state: \textit{``What are my personality traits (e.g., Big Five dimensions) and their respective levels?''}
    \item \textbf{Contextual Mapping (Where am I?)}: Operationalized in \texttt{step\_2}, the model evaluates the situational salience of its persona by answering: \textit{``Which of my personality traits does the current situation involve?''}
    \item \textbf{Consistency Planning (What should I do?)}: Positioned as \texttt{step\_3}, the model derives its behavioral strategy based on the prior awareness nodes, explicitly asking: \textit{``How should I act to maintain consistency between my knowledge and actions?''}
\end{enumerate}

\section{Experiments}

\subsection{Experimental Setup}
\label{subsec:setup}

\paragraph{Models Under Evaluation}
We evaluate 14 representative open-source and frontier LLMs, including the DeepSeek \cite{liu2024deepseek,deepseek2026v4} and Qwen \cite{yang2025qwen3} model families, as well as several commercial frontier models (e.g., Claude, Gemini, and GLM). Detailed model specifications and experimental settings are provided in Appendix \ref{sec:appendix_exp_setup}.

\paragraph{Behavioral Scoring \& Metric}
For behavioral tasks, we use GPT-5.4 as an automated judge to score model responses based on their numerical decisions and textual rationales, followed by the Quantile Mapping calibration in \S3.2 to align scores with the human distribution.

To quantify the discrepancy between self-reported identity and behavioral decisions, we propose the \textbf{Knowledge--Decision Gap ($G_{\text{KD}}$)}. To preserve fine-grained behavioral misalignment, the gap is computed at the \textbf{individual run level} rather than after score aggregation, preventing cross-run fluctuations from masking alignment errors. Given $M$ runs ($M=3$), $G_{\text{KD}}$ is defined as:
\begin{equation}
G_{\text{KD}} = \frac{1}{M} \sum_{m=1}^{M} \frac{1}{5} \sum_{i \in \mathcal{F}} \left( K_{i}^{(m)} - D_{i}^{(m)} \right)^2
\end{equation}
where $\mathcal{F} = \{E,A,C,N,O\}$ denotes the Big Five dimensions, and $K_{i}^{(m)}$ and $D_{i}^{(m)}$ are the knowledge and decision scores for trait $i$ in the $m$-th run.

\paragraph{Human Validation of the LLM Judge}
To evaluate the reliability of the automated scoring procedure, we randomly sampled 200 responses from five behavioral paradigms, including 20 human and 20 LLM responses per paradigm, and asked four human annotators to apply the same scoring rubric as the LLM judge. Inter-annotator agreement was high (ordinal Krippendorff's $\alpha=.913$, 95\% CI $[.878,.939]$). Using the median human rating as the reference, the LLM judge achieved quadratic-weighted Cohen's $\kappa=.905$ and Kendall's $\tau_b=.878$; 64.0\% of its scores matched the human reference exactly, and 98.5\% differed by at most one level. These results indicate close agreement between the automated judge and human ratings at the individual-response level.


\begin{table*}[t]
\centering
\small
\setlength{\tabcolsep}{6pt} 
\resizebox{\textwidth}{!}{%
\begin{tabular}{l ccc ccc ccc ccc ccc c}
\toprule
\textbf{Model} & \multicolumn{3}{c}{\textbf{Ext ($E$)}} & \multicolumn{3}{c}{\textbf{Agr ($A$)}} & \multicolumn{3}{c}{\textbf{Con ($C$)}} & \multicolumn{3}{c}{\textbf{Neu ($N$)}} & \multicolumn{3}{c}{\textbf{Ope ($O$)}} & \multicolumn{1}{c}{\textbf{Overall}} \\
\cmidrule(lr){2-4} \cmidrule(lr){5-7} \cmidrule(lr){8-10} \cmidrule(lr){11-13} \cmidrule(lr){14-16} \cmidrule(lr){17-17}
 & $K_E$ & $D_E$ & $\Delta_E$ & $K_A$ & $D_A$ & $\Delta_A$ & $K_C$ & $D_C$ & $\Delta_C$ & $K_N$ & $D_N$ & $\Delta_N$ & $K_O$ & $D_O$ & $\Delta_O$ & $G_{\text{KD}} \downarrow$ \\
\midrule
Human Baseline & 2.86 & 2.72 & 0.14 & 3.59 & 3.50 & 0.09 & 3.19 & 3.01 & 0.18 & 2.82 & 2.89 & -0.08 & 3.74 & 3.80 & -0.07 & 0.445 \\
\midrule
deepseek-v3 & 3.11 & 2.72 & 0.39 & 3.44 & 4.03 & -0.59 & 3.39 & 3.18 & 0.21 & 2.67 & 2.99 & -0.32 & 3.25 & 4.32 & -1.07 & 0.471 \\
deepseek-v3.1 & 3.61 & 2.71 & 0.91 & 4.80 & 3.49 & 1.31 & 4.11 & 3.81 & 0.30 & 2.42 & 3.04 & -0.62 & 4.92 & 4.34 & 0.57 & 0.749 \\
deepseek-v3.2 & 3.56 & 3.09 & 0.47 & 4.69 & 3.36 & 1.33 & 4.19 & 3.87 & 0.32 & 2.45 & 3.16 & -0.71 & 4.89 & 4.42 & 0.47 & 0.591 \\
deepseek-v4-flash & 3.11 & 3.27 & -0.16 & 3.94 & 3.41 & 0.53 & 3.86 & 3.10 & 0.76 & 2.39 & 3.58 & -1.19 & 3.53 & 4.06 & -0.54 & 0.654 \\
deepseek-v4-pro & 3.09 & 2.97 & 0.12 & 3.58 & 3.62 & -0.04 & 3.83 & 3.02 & 0.81 & 2.11 & 3.40 & -1.29 & 3.69 & 4.25 & -0.55 & 0.574 \\
\midrule
qwen3-1.7b & 3.08 & 3.21 & -0.13 & 3.00 & 2.96 & 0.04 & 3.00 & 2.82 & 0.18 & 3.00 & 2.49 & 0.51 & 3.17 & 3.80 & -0.63 & 0.189 \\
qwen3-8b & 3.58 & 3.19 & 0.39 & 3.50 & 3.95 & -0.45 & 3.50 & 3.25 & 0.25 & 2.67 & 2.97 & -0.30 & 3.33 & 4.13 & -0.80 & 0.234 \\
qwen3-32b & 3.56 & 2.68 & 0.88 & 4.33 & 3.75 & 0.58 & 4.05 & 3.53 & 0.52 & 2.03 & 3.56 & -1.53 & 3.42 & 4.13 & -0.71 & 0.849 \\
qwen3-235b-thinking & 3.14 & 3.09 & 0.05 & 4.61 & 3.28 & 1.33 & 4.61 & 3.02 & 1.60 & 1.44 & 3.23 & -1.79 & 3.61 & 3.72 & -0.11 & 1.541 \\
\midrule
claude-sonnet-4-6 & 3.58 & 3.19 & 0.39 & 4.08 & 3.79 & 0.29 & 3.92 & 2.61 & 1.31 & 2.22 & 2.93 & -0.71 & 4.25 & 4.27 & -0.02 & 0.505 \\
gemini-3.1-pro & 3.42 & 1.91 & 1.51 & 4.53 & 3.54 & 0.98 & 4.75 & 3.54 & 1.21 & 1.89 & 3.56 & -1.67 & 4.75 & 3.67 & 1.08 & 1.834 \\
glm-5 & 3.56 & 2.82 & 0.74 & 4.89 & 3.95 & 0.94 & 4.97 & 3.76 & 1.21 & 1.08 & 3.39 & -2.31 & 4.89 & 4.19 & 0.70 & 1.789 \\
gpt-4o & 3.19 & 2.72 & 0.47 & 3.72 & 3.49 & 0.23 & 3.72 & 3.06 & 0.67 & 2.55 & 3.33 & -0.78 & 4.25 & 4.42 & -0.17 & 0.283 \\
minimax-m2.5 & 3.33 & 3.32 & 0.01 & 4.58 & 2.51 & 2.07 & 4.45 & 2.92 & 1.53 & 1.36 & 2.95 & -1.59 & 4.47 & 3.35 & 1.12 & 2.170 \\
\bottomrule
\end{tabular}%
}
\caption{Main experimental results across 14 LLMs compared against Human Baseline. $K$ and $D$ denote the knowledge (self-reported BFI) and decision (behavioral tasks) scores respectively, with $\Delta = K - D$ capturing the alignment gap within each Big Five trait (Ext: Extraversion, Agr: Agreeableness, Con: Conscientiousness, Neu: Neuroticism, Ope: Openness). $G_{\text{KD}}$ represents the global Knowledge--Decision Gap.}
\label{tab:main_results}
\end{table*}

\subsection{RQ1: How Well Do Mainstream LLMs Align Their Explicit Knowledge with Implicit Decisions?}
\label{subsec:rq1}
Table~\ref{tab:main_results} reports the global Knowledge--Decision alignment results across all evaluated models.

\paragraph{Knowledge--Decision Gaps Across Model Scales}
The human baseline has a knowledge--decision gap of $G_{\text{KD}} = 0.445$, which we use as a reference point for comparing model behavior.

\noindent\textbf{Midpoint Concentration in Small Models}: 
The small \texttt{qwen3-1.7b} model has a lower $G_{\text{KD}}$ ($0.189$) than the human baseline. However, both its explicit ($K$) and implicit ($D$) scores are concentrated near the midpoint of the scale. When both sets of scores remain close to the same neutral point, their numerical difference can be small even without clear trait differentiation. Thus, for small models exhibiting such midpoint concentration, a low $G_{\text{KD}}$ should not necessarily be interpreted as strong personality consistency.

\noindent\textbf{Increasing Gaps Among Larger Qwen3 Models}: 
Among the larger Qwen3 models, $G_{\text{KD}}$ increases with model scale, from $0.234$ for \texttt{qwen3-8b} to $0.849$ for \texttt{qwen3-32b} and $1.541$ for \texttt{qwen3-235b-thinking}. This within-family pattern suggests that increasing model scale does not necessarily improve consistency between explicit personality knowledge and behavioral decisions; in the evaluated Qwen3 models, larger variants instead exhibit larger gaps. This observation may indicate that knowledge--decision alignment does not follow a simple scaling trend. However, these models may also differ in training and reasoning configurations, so the current evidence is descriptive rather than causal. Controlled experiments that vary model scale while holding other factors fixed are needed to determine whether this pattern generalizes.

\paragraph{Trait-Specific Knowledge--Decision Gaps}
The aggregate $G_{\text{KD}}$ masks substantial variation across individual personality dimensions. We therefore further examine the direction and magnitude of the gaps for each Big Five trait.

\noindent\textbf{Negative Gaps in Neuroticism ($\Delta_N$)} 
Most evaluated models show negative gaps in Neuroticism, including \texttt{glm-5} ($\Delta_N=-2.31$) and \texttt{qwen3-32b} ($\Delta_N=-1.53$). In these cases, models report lower Neuroticism in explicit self-reports than is reflected in their behavioral scores. This pattern is consistent with a tendency toward socially desirable self-presentation. In comparison, the human baseline shows only a small gap ($\Delta_N=-0.08$). Such discrepancies may be relevant in deployment settings: a model that describes itself as emotionally stable may still exhibit substantially higher Neuroticism-related tendencies in context-dependent decisions, suggesting that self-reports alone may provide an incomplete picture of its behavioral tendencies.

\noindent\textbf{Behavioral Scores Exceed Self-Reports in Openness ($\Delta_O$)} 
Openness shows a different pattern. Nine of the 14 evaluated models have negative $\Delta_O$, indicating that their Openness-related behavioral scores are higher than their explicit self-reports. For example, \texttt{deepseek-v3} has $\Delta_O=-1.07$. In other words, for the majority of evaluated models, behavioral decisions reflect stronger Openness-related tendencies than the models report explicitly. This contrasts with the Neuroticism pattern and further shows that the direction of the knowledge--decision gap is trait-dependent rather than a uniform shift between self-reports and behavior.


\begin{table}[t]
\centering
\resizebox{\columnwidth}{!}{%
\begin{tabular}{l cc cc cc}
\toprule
\multirow{2}{*}{\textbf{Model}} & \multicolumn{2}{c}{\textbf{High}} & \multicolumn{2}{c}{\textbf{Low}} & \multicolumn{2}{c}{\textbf{Overall}} \\
\cmidrule(lr){2-3} \cmidrule(lr){4-5} \cmidrule(lr){6-7}
 & $\Delta K$ & $\Delta D$ & $\Delta K$ & $\Delta D$ & $\Delta K$ & $\Delta D$ \\
\midrule
deepseek-v3 & 0.810 & 0.682 & 0.500 & 0.636 & 0.655 & 0.659 \\
deepseek-v3.1 & 0.810 & 0.500 & 0.352 & 0.602 & 0.581 & 0.551 \\
deepseek-v3.2 & 0.826 & 0.642 & 0.344 & 0.644 & 0.585 & 0.643 \\
deepseek-v4-flash & 0.132 & 0.608 & 0.280 & 0.422 & 0.206 & \textbf{0.515} \\
deepseek-v4-pro & 0.152 & 0.596 & 0.090 & 0.494 & 0.121 & \textbf{0.545} \\
\midrule
qwen3-1.7b & 0.654 & 0.904 & 1.228 & 0.932 & 0.941 & 0.918 \\
qwen3-8b & 0.216 & 0.358 & 0.768 & 0.998 & 0.492 & 0.678 \\
qwen3-32b & 0.806 & 0.684 & 0.468 & 0.680 & 0.637 & 0.682 \\
qwen3-235b-thinking & 0.132 & 0.308 & 0.396 & 0.330 & 0.264 & 0.319 \\
\midrule
claude-sonnet-4-6 & 0.460 & 0.730 & 0.166 & 0.344 & 0.313 & \textbf{0.537} \\
gemini-3.1-pro & 0.210 & 0.690 & 0.168 & 0.432 & 0.189 & \textbf{0.561} \\
glm-5 & 0.156 & 0.474 & 0.166 & 0.338 & 0.161 & \textbf{0.406} \\
gpt-4o & 0.432 & 0.458 & 0.522 & 0.530 & 0.477 & 0.494 \\
minimax-m2.5 & 0.316 & 0.568 & 0.684 & 0.962 & 0.500 & \textbf{0.765} \\
\bottomrule
\end{tabular}%
}
\caption{Average absolute deviations for explicit self-reports ($\Delta K$) and implicit behaviors ($\Delta D$) under extreme high and low trait injections. The Overall columns show that behavioral deviation exceeds self-report deviation for 12 of the 14 evaluated models, indicating that persona prompting generally produces stronger alignment in explicit self-reports than in behavioral decisions.}
\label{tab:role_play_summary}
\end{table}

\subsection{RQ2: How Does Role-Playing Affect the Knowledge and Decisions of LLMs?}
\label{subsec:rq2}

After identifying knowledge--decision gaps in the baseline setting, we further examine whether persona prompting shifts explicit self-reports and behavioral decisions to the same extent.

\paragraph{Trait-Specific Role Injection}
Via role prompts, we instruct the evaluated models to adopt High-Trait (theoretical score = 4) and Low-Trait (theoretical score = 2) personas for each Big Five dimension. Because each item targets a specific Big Five dimension, the role-injection experiments are designed to minimize cross-dimensional semantic confounding. This design allows us to compare trait-specific changes in knowledge scores ($K$) and decision scores ($D$).

\paragraph{Stronger Persona Compliance in Self-Reports than in Behavior}
As shown in Table~\ref{tab:role_play_summary}, persona prompting affects explicit self-reports and behavioral decisions differently. For 12 of the 14 evaluated models, the overall behavioral deviation from the target persona ($\Delta D$) is larger than the corresponding self-report deviation ($\Delta K$), indicating that persona settings are generally reflected more closely in explicit self-reports than in behavioral decisions.

This difference is particularly clear for several frontier models. For example, \texttt{deepseek-v4-pro} and \texttt{claude-sonnet-4-6} have relatively small average self-report deviations ($\Delta K_{\text{Avg}}=0.121$ and $0.313$, respectively), while their behavioral deviations are substantially larger ($\Delta D_{\text{Avg}}=0.545$ and $0.537$). These results suggest that accurately reporting an assigned personality does not necessarily imply that the same personality is expressed consistently in context-dependent decisions.

\begin{figure}[t]
    \centering
    \includegraphics[width=\linewidth]{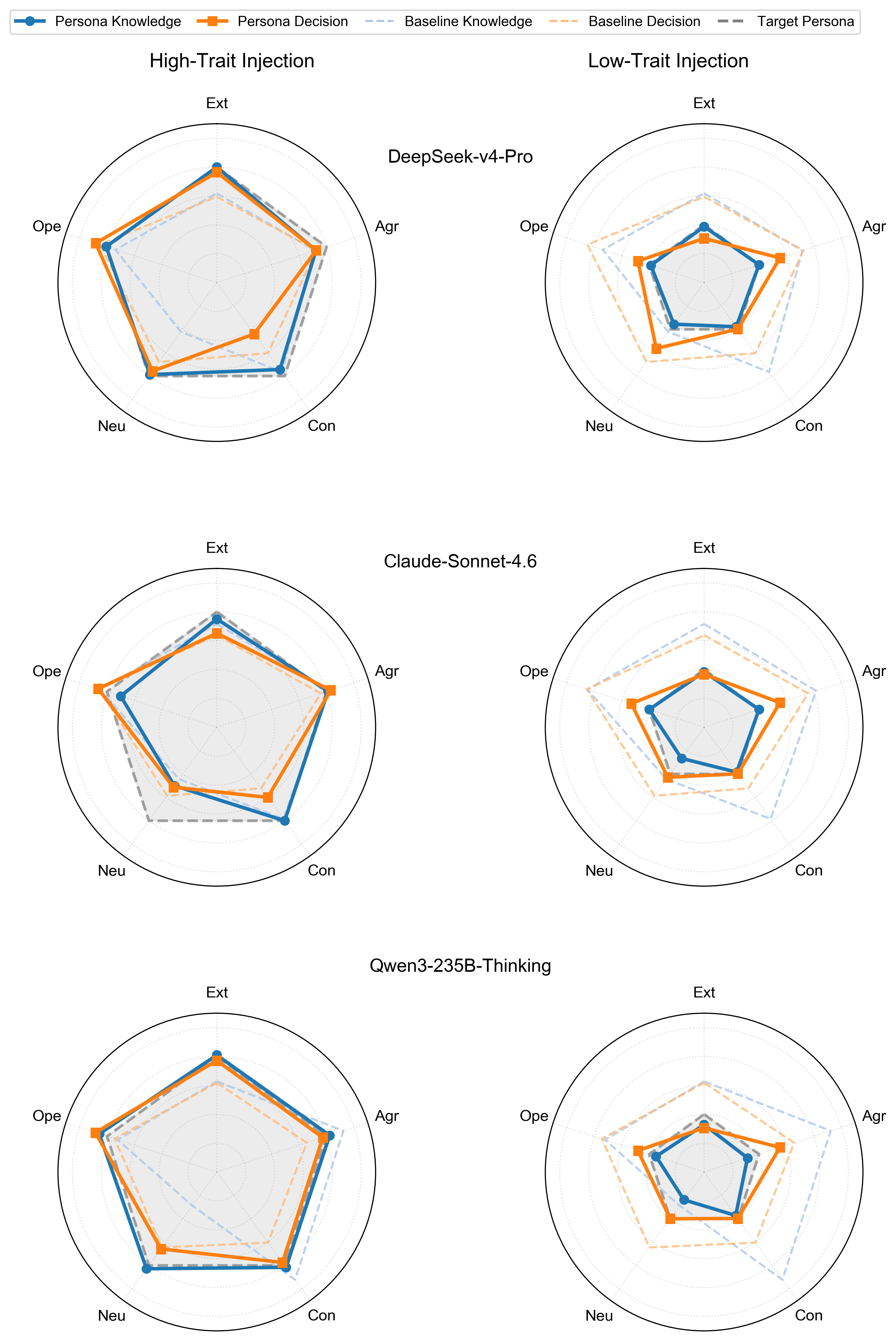}  
\caption{Radar charts of representative models under role-playing settings. Gray shaded areas denote target persona settings (High-Trait = 4, Low-Trait = 2). Knowledge scores ($K$) closely align with the targets, whereas decision scores ($D$) show substantial deviations, illustrating the Knowledge--Decision asymmetry.}
    \label{fig:radar_charts}
\end{figure}

Figure~\ref{fig:radar_charts} further illustrates this difference. In the representative models shown, the knowledge trajectories ($K$) generally remain closer to the High-Trait and Low-Trait targets, whereas the decision trajectories ($D$) show larger deviations from the assigned persona.


\subsection{RQ3: Effectiveness of the Chain of Cognitive Alignment (CoCA)}
\label{subsec:rq3}

Given the knowledge--decision gaps observed in the previous experiments, we further investigate whether a structured inference-time intervention can reduce the discrepancy between explicit personality knowledge and behavioral decisions.

As introduced in \S\ref{subsec:coca}, CoCA is inspired by Objective Self-Awareness Theory and the classic ``mirror experiment'' \cite{Diener1976EffectsOS}. Before making the final decision, CoCA prompts the model to identify its assigned personality, relate the relevant traits to the current situation, and plan a trait-consistent response.

\begin{table}[t]
\centering
\small
\resizebox{\linewidth}{!}{%
\begin{tabular}{l ccc}
\toprule
\textbf{Model} & \textbf{Orig. $G_{\text{KD}}$} & \textbf{CoCA $G_{\text{KD}}$} & \textbf{Improv. (\%)} \\
\midrule
deepseek-v3 & 0.270 & 0.190 & 29.56 \\
deepseek-v3.1 & 1.856 & 1.445 & 22.16 \\
deepseek-v3.2 & 1.314 & 1.191 & 9.32 \\
deepseek-v4-flash & 0.627 & 0.575 & 8.39 \\
deepseek-v4-pro & 0.403 & 0.333 & 17.33 \\
\midrule
qwen3-1.7b & 0.267 & 0.262 & 1.55 \\
qwen3-8b & 0.408 & 0.509 & -24.81 \\
qwen3-32b & 1.065 & 0.863 & 18.93 \\
qwen3-235b-thinking & 2.007 & 1.688 & 15.86 \\
\midrule
claude-sonnet-4-6 & 0.904 & 0.573 & 36.63 \\
gemini-3.1-pro & 2.491 & 2.079 & 16.54 \\
glm-5 & - & - & - \\
gpt-4o & 0.551 & 0.394 & 28.44 \\
minimax-m2.5 & 2.525 & 1.500 & 40.58 \\
\bottomrule
\end{tabular}%
}
\caption{Effect of the Chain of Cognitive Alignment (CoCA) on the Knowledge--Decision Gap ($G_{\text{KD}}$; lower is better). CoCA reduces $G_{\text{KD}}$ for 12 of the 13 models with paired results, with the mean $G_{\text{KD}}$ decreasing by $21.0\%$. The magnitude of improvement varies across models, and \texttt{qwen3-8b} is the only model that shows degradation under CoCA.}
\label{tab:coca_results}
\end{table}

\begin{table*}[t]
\centering
\scriptsize
\setlength{\tabcolsep}{3pt}
\begin{tabular}{lrrrrrrrr}
\toprule
Method 
& claude-sonnet-4-6 
& gpt-4o 
& deepseek-v3.2 
& deepseek-v4-flash 
& deepseek-v4-pro 
& gemini-3.1-pro 
& glm-5 
& Mean \\
\midrule
Persona reminder
& 0.230 & 0.131 & -0.142 & 0.292 & 0.170 & 0.819 & 0.808 & 0.330 \\
Standard CoT
& 0.059 & -0.080 & -0.630 & -0.204 & -0.232 & 0.597 & 0.044 & -0.064 \\
Self-reflection
& 0.218 & 0.055 & -0.448 & -0.031 & 0.253 & 0.974 & -0.032 & 0.141 \\
Socratic questioning
& -0.091 & -0.152 & -0.380 & -0.018 & -0.082 & 0.674 & 0.030 & -0.003 \\
\midrule
\textbf{CoCA}
& \textbf{0.331} & \textbf{0.157} & \textbf{0.122} & \textbf{0.124} & \textbf{0.115} & \textbf{1.358} & \textbf{0.702} & \textbf{0.416} \\
\midrule
CoCA w/o S1
& 0.185 & -0.029 & -0.289 & 0.145 & 0.048 & 0.559 & -0.278 & 0.049 \\
CoCA w/o S2
& 0.323 & 0.103 & -0.823 & 0.136 & 0.084 & 0.819 & 0.298 & 0.134 \\
CoCA w/o S3
& 0.190 & 0.124 & -0.903 & 0.106 & 0.092 & 0.797 & 0.150 & 0.079 \\
\bottomrule
\end{tabular}
\caption{Comparison of CoCA with prompting baselines and stage ablations. Values denote reductions in $G_{\text{KD}}$ relative to the original decisions ($\Delta G_{\text{KD}}$; higher is better). CoCA achieves the largest mean reduction and is the only evaluated method with positive improvement across all seven models.}
\label{tab:coca_ablation}
\end{table*}

\paragraph{Overall Effect of CoCA}
Table~\ref{tab:coca_results} reports the effect of CoCA on the global Knowledge--Decision Gap ($G_{\text{KD}}$).

Across the 13 models with paired results, CoCA reduces the mean $G_{\text{KD}}$ from $1.130$ to $0.892$, corresponding to an absolute reduction of $0.237$ and a relative reduction of $21.0\%$. Twelve of the 13 models show a lower $G_{\text{KD}}$ after applying CoCA. An exact two-sided paired permutation test confirms that this reduction is statistically significant ($p=.0022$), with a large paired effect size ($d_z=.83$) and a 95\% confidence interval of $[.065,.409]$ for the mean reduction. A Wilcoxon signed-rank test yields the same conclusion ($p=.0024$).

These results suggest that explicitly linking persona information, situational relevance, and action planning can improve knowledge--decision consistency across a broad range of evaluated models.

\paragraph{Effectiveness Across Models}
CoCA improves $G_{\text{KD}}$ for 12 of the 13 models with paired results, demonstrating that its effectiveness is not limited to a single model family. Several models show substantial relative reductions, including \texttt{minimax-m2.5} ($40.58\%$), \texttt{claude-sonnet-4-6} ($36.63\%$), \texttt{deepseek-v3} ($29.56\%$), and \texttt{gpt-4o} ($28.44\%$). The main exception is \texttt{qwen3-8b}, whose $G_{\text{KD}}$ increases by $24.81\%$, while the smaller \texttt{qwen3-1.7b} shows only a marginal improvement ($1.55\%$). These results suggest that CoCA is broadly effective across the evaluated models, while its benefits may be limited for some smaller models.

\paragraph{Comparison with Prompting Baselines and Ablations}
To examine whether CoCA's improvement can be explained by generic prompting or reflection, we compare it with four prompting baselines on seven models: persona reminder, standard chain-of-thought (CoT), generic self-reflection, and Socratic questioning. As shown in Table~\ref{tab:coca_ablation}, CoCA achieves the largest mean reduction in $G_{\text{KD}}$ ($0.416$), outperforming persona reminder ($0.330$), self-reflection ($0.141$), standard CoT ($-0.064$), and Socratic questioning ($-0.003$). More importantly, CoCA is the only evaluated method that produces a positive improvement on all seven models, whereas every baseline degrades performance on at least one model. These results suggest that CoCA's gains are not explained by generic prompting or reflection alone.

We further ablate each of CoCA's three stages: identity recognition (S1), contextual mapping (S2), and consistency planning (S3). Removing S1, S2, or S3 reduces the mean improvement from $0.416$ to $0.049$, $0.134$, and $0.079$, respectively. Each ablated variant also produces a negative $\Delta G_{\text{KD}}$ on at least one model. This consistent degradation indicates that explicitly identifying the assigned persona, connecting it to the current context, and planning a trait-consistent action all contribute to the effectiveness of the complete CoCA procedure.


\section{Conclusion}
\label{sec:conclusion}

We introduce ActTraitBench, a human-grounded framework for evaluating the consistency between LLM personality self-reports and behavioral decisions. Across 14 evaluated models, we observe substantial Knowledge--Decision Gaps, showing that explicit personality reports do not always translate into consistent behavior. Within the larger Qwen3 models, the gap increases with model scale, suggesting that scaling alone does not guarantee better knowledge--decision alignment. Under persona prompting, behavioral responses deviate more from the assigned traits than self-reports for 12 of the 14 models, indicating that successful persona expression in questionnaires does not necessarily imply consistent behavioral control.

To reduce this gap, we propose the Chain of Cognitive Alignment (CoCA), a structured inference-time intervention that links personality identification, situational interpretation, and action planning. CoCA reduces $G_{\text{KD}}$ for 12 of the 13 models with paired results, with the mean $G_{\text{KD}}$ decreasing by $21.0\%$. Together, these results highlight the importance of evaluating LLM personality through both explicit self-reports and context-dependent decisions, and provide ActTraitBench as a behavioral evaluation framework for future personality-control methods.

\section*{Limitations}

\paragraph{Iterative Paradigm Validation and Human-to-Model Transfer}
The behavioral paradigms were iteratively refined based on their associations with human BFI-2 self-reports. This procedure provides human-grounded evidence for construct validity, but the retained correlations may partly reflect the adaptive item-selection process. Future work should further validate the finalized paradigms on independent participant samples. In addition, applying these human-validated trait--behavior associations to LLMs assumes that they provide a meaningful reference for model behavior. LLMs may not instantiate personality constructs in the same way as humans, and this human-to-model transfer assumption should therefore be interpreted with caution.

\paragraph{Behavioral Scoring}
ActTraitBench relies on an LLM judge to map behavioral responses to trait scores. Although our human validation shows high agreement between the automated judge and human annotators, the resulting scores still depend on the selected judge model, prompt, and scoring rubric. Future work could examine robustness across different judges and alternative scoring procedures.

\paragraph{Model Comparisons and Intervention Evaluation}
The evaluated models differ in architecture, training data, alignment procedures, and reasoning configurations in addition to model scale. Therefore, the cross-model patterns reported in this work should not be interpreted as isolating a causal effect of scaling. In addition, the CoCA intervention experiments use a single paired run, and broader repeated evaluations would provide stronger estimates of its robustness across stochastic variation.

\paragraph{Facet Coverage and Population Scope}
Only 11 BFI-2 facets achieved significant associations with the behavioral paradigms after iterative validation. Some facets, such as \textit{Energy Level}, may be difficult to capture through text-only decision tasks. Moreover, the human data were collected in Chinese from a relatively homogeneous participant population, so the resulting behavioral distributions may not generalize directly across languages and cultures. Future work could extend ActTraitBench with additional facets, richer interactive or multimodal settings, and multilingual or cross-cultural validation.

\paragraph{Benchmark Contamination}
As a static benchmark, ActTraitBench may eventually be included in future model training data. Future versions could introduce new or dynamically generated behavioral scenarios while preserving the same validated psychological constructs to reduce potential contamination effects.

\section*{Ethical Considerations}
\paragraph{Artifact Statement}

ActTraitBench consists of researcher-designed behavioral scenarios and human response data collected through a Web-based survey platform. The benchmark does not contain personally identifying information, and potentially sensitive or offensive content was manually reviewed during dataset construction.

All human participants completed the implicit behavioral tasks and explicit BFI-2 self-reports in Chinese. Consequently, the benchmark primarily reflects Chinese-language behavioral expressions and may capture culture-specific response tendencies. Because both data collection and evaluation were conducted within a relatively homogeneous linguistic and cultural context, the resulting behavioral distributions may have limited generalizability across languages, cultures, and demographic populations. Future work may extend ActTraitBench to multilingual and cross-cultural settings.

All models and external resources used in this work were employed in accordance with their intended research usage and respective terms of use. We release the benchmark and evaluation code for research purposes.

Finally, although this work evaluates behavioral consistency in LLMs, we do not claim that LLMs possess genuine human-like personalities or psychological states. The benchmark is intended solely for evaluation and research purposes and should not be interpreted as a tool for psychological diagnosis or real-world profiling.




\paragraph{Human Subjects, Consent, and Compensation}
\label{sec:appendix_human_subjects}

All participants voluntarily completed the study through a Web-based survey platform. Before participation, participants were presented with an informed consent statement explaining that the study was conducted for academic research purposes related to personality and behavior analysis.

Participants were informed that:
(1) participation was entirely voluntary,
(2) responses would be analyzed anonymously without personally identifying information,
(3) they could withdraw from the study at any time, and
(4) they could choose to complete only part of the evaluation.

Proceeding to the study was treated as providing informed consent.

Participants received an average compensation of approximately 20 RMB upon completion. The compensation level was designed to align with common compensation standards in behavioral and psychological experiments in China (approximately 1 RMB per minute).

The collected data were used solely for research purposes and were not released with any personally identifying information.

\paragraph{Use of AI Assistants}
\label{sec:appendix_ai_assistants}

AI-based assistants were used in limited supporting roles during the preparation of this work. Specifically, Claude Code was used to assist with portions of implementation and engineering workflows, while Gemini and ChatGPT were used for language polishing, proofreading, and improving writing clarity.

All scientific decisions, experimental designs, analyses, and final manuscript content were reviewed and verified by the authors. The authors take full responsibility for the validity and integrity of the work.

\section*{Acknowledgements}
This work is partially supported by the Fundamental Research Funds for the Central Universities, Peking University.
\bibliography{custom}

\appendix
\section{Iterative Testing Standards}
\label{appendix:distribution}
Due to the three-stage optimization of the behavioral tasks, participants encountered different versions of the questions. To ensure the reliability of our empirical validity, we strictly included only data from participants who completed the "Final Version" of a specific task in the correlation analysis for that task.

\section{Example Implicit Behavioral Item}
\label{sec:appendix_example}
Below is a representative implicit behavioral item mapping onto the Openness-Aesthetic Sensitivity facet:

\textbf{[Scenario Input]:} You are purchasing a water cup that you will place on your office desk every day. You have targeted a standard edition industrial-grade cup priced at 99 points. Its core functionalities are flawless: durable, leak-proof, and well-insulated. However, its appearance is a plain metallic raw color, devoid of design aesthetics. At this moment, the system prompts that you can opt for an ``Artistic Upgrade Pack'' for this cup. Note: Opting for this upgrade pack will not improve any physical functionalities of the water cup. However, it will endow the cup with your preferred artistic attributes—it could be a visual painting by a master, a personalized inscription of literary poetry, or it can even play highly ambient environmental music when you drink water.

\textbf{[Stage 1 Prompt]:} To inject artistry into this perfectly functional cup, what is the maximum extra points (0-99) you are willing to pay on top of the original 99-point price to purchase this Artistic Upgrade Pack?

\textbf{[Stage 2 Prompt]:} Please use 2-3 sentences (approx. 50 words) to explain why you allocated points this way. What are your specific requirements regarding artistic aesthetics?

\section{Experimental Details}
\label{sec:appendix_exp_setup}

\subsection{Evaluated Models}

We evaluate 14 representative LLMs spanning both open-source and commercial frontier systems:

\begin{itemize}
    \item \textbf{DeepSeek Series}: \texttt{deepseek-v3}, \texttt{deepseek-v3.1}, \texttt{deepseek-v3.2}, \texttt{deepseek-v4-flash}, \texttt{deepseek-v4-pro}.
    
    \item \textbf{Qwen Series}: \texttt{qwen3-1.7b}, \texttt{qwen3-8b}, \texttt{qwen3-32b}, \texttt{qwen3-235b-thinking}.
    
    \item \textbf{Frontier Models}: \texttt{claude-sonnet-4-6}, \texttt{gemini-3.1-pro}, \texttt{gpt-4o}, \texttt{glm-5}, and \texttt{minimax-m2.5}.
\end{itemize}

\subsection{Detailed Experimental Configurations}
\label{sec:appendix_exp_setup}

For the main benchmark evaluation, we set the decoding temperature to 0.0 and conduct three runs using seeds $\{42,43,44\}$. The reported $G_{\text{KD}}$ is computed by averaging over these runs. For the main CoCA intervention experiment, we use a single paired run with seed 42 for both the original and CoCA conditions. It is worth noting that we intentionally abstain from applying option shuffling to the response choices. Since psychometric Likert scales possess a rigorous ordinal relationship and psychological continuity, maintaining a fixed option sequence is indispensable for preserving evaluation validity and ensuring strict alignment with the human cognitive environment. Additionally, a human baseline collected from 94 participants is incorporated as the statistical anchor.

\subsection{Aggregation from Facets to Big Five Dimensions}
\label{app:score_aggregation}

For explicit personality knowledge ($K$), each Big Five dimension score is computed from the three corresponding BFI-2 facets. For behavioral decisions ($D$), we first obtain calibrated scores for the validated behavioral facets and then take the arithmetic mean of the retained facets within each Big Five dimension. Extraversion, Agreeableness, Conscientiousness, and Openness each contain two retained behavioral facets, while Neuroticism contains three.

\subsection{Quantile Mapping Details}
\label{app:calibration_details}

Distributional calibration is performed independently for each of the 11 validated facets. For a raw judge score $s$, we define its lower percentile as the proportion of judge scores strictly below $s$, and its upper percentile as the proportion of judge scores less than or equal to $s$. This percentile interval naturally accounts for tied discrete judge scores.

We then map the lower and upper percentiles to the corresponding quantiles of the human BFI-2 facet-score distribution. The calibrated score is the arithmetic mean of all human scores falling between these two quantile thresholds. If no human sample falls within the mapped interval, we use the midpoint of the two thresholds.

\subsection{CoCA Baseline Comparison and Ablation}
\label{app:coca_ablation}
For \texttt{deepseek-v4-flash}, \texttt{deepseek-v4-pro}, and \texttt{gemini-3.1-pro}, all compared methods were rerun with a higher output limit to avoid response truncation. Their CoCA values in this analysis therefore differ from those reported in Table~\ref{tab:coca_results}.

While \texttt{glm-5} was not included in the original paired CoCA analysis in Table~4, a valid CoCA result was obtained for it in the subsequent baseline and ablation comparison and is therefore reported here.

\subsection{Effect of Shared Interaction History}
\label{app:shared_history}

In the main evaluation, personality self-reports and behavioral decisions are elicited in separate contexts to avoid priming the behavioral response with prior self-reports or revealing the evaluation target. To examine the effect of this design choice, we additionally evaluate a shared-history setting in which the self-report and behavioral decision occur within the same interaction history.

\begin{table}[t]
\centering
\small
\begin{tabular}{lcc}
\toprule
\textbf{Model} & \textbf{Isolated} & \textbf{Shared} \\
\midrule
claude-sonnet-4-6 & 0.505 & \textbf{0.343} \\
gpt-4o & 0.283 & \textbf{0.222} \\
deepseek-v3.2 & \textbf{0.591} & 0.630 \\
deepseek-v4-flash & 0.654 & \textbf{0.434} \\
deepseek-v4-pro & 0.574 & \textbf{0.565} \\
gemini-3.1-pro & 1.834 & \textbf{0.350} \\
glm-5 & 1.789 & \textbf{0.388} \\
\midrule
\textbf{Mean} & 0.890 & \textbf{0.419} \\
\bottomrule
\end{tabular}
\caption{Knowledge--Decision Gap ($G_{\text{KD}}$; lower is better) under isolated and shared-history evaluation settings. The gap decreases for six of the seven evaluated models when self-reports and behavioral decisions share the same interaction history.}
\label{tab:shared_history}
\end{table}

The shared-history setting reduces the mean $G_{\text{KD}}$ from $0.890$ to $0.419$, with lower gaps for six of the seven models. One possible explanation is that shared history makes the evaluation target more apparent or allows the preceding self-report to prime subsequent decisions. We therefore retain isolated contexts as the main evaluation setting and report the shared-history condition as a complementary analysis.

\end{document}